\documentclass{ecai}  

\usepackage{graphicx}
\usepackage{latexsym}

\usepackage{amsmath,amssymb,amsfonts}
\usepackage[linguistics]{forest}
\usepackage{url}
\usepackage{caption}
\usepackage{subcaption}
\usepackage{xcolor}
\usepackage{multirow}
\usepackage[section]{placeins}

\begin{document}

\begin{frontmatter}

\title{Forecast reconciliation for vaccine supply chain optimization}

\author[A]{\fnms{Bhanu}~\snm{Angam}}
\author[B]{\fnms{Alessandro}~\snm{Beretta}}
\author[A]{\fnms{Eli}~\snm{De Poorter}\orcid{0000-0002-0214-5751}}
\author[B]{\fnms{Matthieu}~\snm{Duvinage}}
\author[A]{\fnms{Daniel}~\snm{Peralta}\orcid{0000-0002-7544-8411}\thanks{Corresponding Author. Email: daniel.peralta@ugent.be.}} 

\address[A]{IDLab, Department of Information Technology, Ghent University - imec, Technologiepark 126, 9052 Gent, Belgium}
\address[B]{GSK, Avenue Fleming, 20, 1300 Wavre - Belgium}

\begin{abstract}
Vaccine supply chain optimization can benefit from hierarchical time series forecasting, when grouping the vaccines by type or location. However, forecasts of different hierarchy levels become incoherent when higher levels do not match the sum of the lower levels forecasts, which can be addressed by reconciliation methods.

In this paper, we tackle the vaccine sale forecasting problem by modeling sales data from GSK between 2010 and 2021 as a hierarchical time series. After forecasting future values with several ARIMA models, we systematically compare the performance of various reconciliation methods, using statistical tests. We also compare the performance of the forecast before and after COVID. The results highlight Minimum Trace and Weighted Least Squares with Structural scaling as the best performing methods, which provided a coherent forecast while reducing the forecast error of the baseline ARIMA.
\end{abstract}

\end{frontmatter}

\section{Introduction}	\label{sec:intro}

Forecasting vaccine demand is critical to practitioners; its underestimation lead to loss of customers and profits, whereas its overestimation would cause the expiry of produced vaccines and a waste of resources~\cite{Andrus2008,Gilchrist2013}. Vaccine demand forecasting is mostly done by replicating values from previous years~\cite{Mueller2016,Cernuschi2018}, sometimes combining them with demographic information~\cite{Debellut2019}. However, machine learning techniques become necessary to obtain better solutions to this problem.

Given enough previous data, vaccine demand can be modeled as a time series, that is, a sequence of values measured in fixed time intervals~\cite{Hyndman2021}. There are various methods to forecast future values in a time series. Among them, linear methods like ARIMA are still widely used, due to their interpretability and stability.
In many domains, time series can be arranged in the form of a hierarchy~\cite{Hyndman2011}. For instance, sales can be split into geographical regions or product types, where the series at one level corresponds to the sum of the values of the series at the immediately lower level. Each of these time series can be independently forecast; however, the coherency of the time series is lost: the forecast of a higher level does not correspond to the sum of the forecasts of lower levels.

Forecast combination methods ensure the coherency of the forecasts by only forecasting the series at one level (typically the highest or the lowest), and then calculating the forecasts for the rest of the hierarchy~\cite{Gross1990,Hollyman2021}. Their main limitation is that they use only one level of the hierarchy to compute the forecasts. Forecasting reconciliation aims at obtaining coherent forecasts based on the entire hierarchy~\cite{Hollyman2021,Hyndman2021,Abolghasemi2022}. Several methods have been proposed in the field, such as Ordinary Least Squares~\cite{Hyndman2011}, Weighted Least Squares~\cite{Hyndman2021} and Trace Minimization~\cite{Wickramasuriya2019}. Frequently, the values of the time series are limited to a certain range, which is often the positive real or integer numbers. Positive outcomes when forecasting are usually enforced by taking the logarithm of the series; however, when applying the reconciliation in the original space it is possible that some forecasts become negative. Therefore, some methods have been proposed to enforce positive reconciled forecasts~\cite{Wickramasuriya2020}.

There are a few works in the scientific literature that model vaccine demand as a time series and then forecast future values. In~\cite{Sahisnu2020}, ARIMA was applied to forecast vaccine demand in Indonesia. In~\cite{Barajas2021}, a cloud infrastructure used a Long Short-Term Memory (LSTM) to forecast the demand for COVID-19 vaccines. However, none of these works consider hierarchical time series to forecast vaccine demand.

In this paper, we compare several state-of-the-art combination and reconciliation methods to adjust the ARIMA forecast of the vaccine sales of a hierarchy of products across several geographical areas. We also evaluate the impact of COVID-19 on the sales, analyzing how the forecast error increases after COVID-19 due to large unexpected changes in demand patterns. We carry out a series of statistical tests on the obtained results, and come to the conclusion that, for this dataset, Minimum Trace and Weighted Least Squares with Structural scaling were the best performing reconciliation methods, achieving reconciled forecasts while also lowering the forecast error.

The main contributions of this paper are the following:
\begin{itemize}
	\item Comparison of various combination and reconciliation methods for adjusting the ARIMA forecast of vaccine sales in a hierarchical product structure.
	\item Evaluation of the impact of COVID-19 on sales and demonstrating an increase in forecast error due to changes in demand patterns.
	\item Finally, through statistical tests, we conclude that Minimum Trace and Weighted Least Squares with Structural scaling are the best performing and most robust reconciliation methods for the dataset used.
\end{itemize} 

This paper is structured as follows. First, Section~\ref{sec:background} introduces the topic of forecast recombination and reconciliation in hierarchical time series. Section~\ref{sec:proposal} describes the methodology carried out in this study, including the dataset analyzed and its preprocessing. The results of the experiments are presented and analyzed in Section~\ref{sec:results}. Finally, Section~\ref{sec:conclusion} presents the conclusions of this study.

\section{Background}	\label{sec:background}

\subsection{Hierarchical time series}	\label{sec:timeseries}
\begin{figure}[ht]
	\begin{center}
		\begin{forest}
			[Total [A [AA] [AB]] [B [BA] [BB] [BC]]]
		\end{forest}  
	\end{center}
	\caption{Example of a simple hierarchy}
	\label{fig:example_hierarchy}
\end{figure}
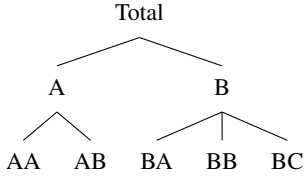

In a hierarchical time series, the values of a particular level can be expressed as the sum of values at the immediately lower level. A hierarchical time series with $n$ observations, $k$ levels, a total of $m$ series, and $m_i$ series in level $i$ can be expressed as a matrix $\mathbf{Y}_t$ with dimensions $m \times n$, where each row contains one of the time series. Therefore, the hierarchical time series can be expressed as $\mathbf{Y}_t=\mathbf{S}\mathbf{Y}_{k,t}$, where $\mathbf{S}$ is a summing matrix with dimensions $m \times m_k$ and $\mathbf{Y}_{k,t}$ contains the series at the bottom level of the hierarchy. The example hierarchy shown in Fig.~\ref{fig:example_hierarchy} can be represented as (\ref{eq:Smatrixexample}). Note that the bottom $m_k$ rows of $\mathbf{S}$ are the identity matrix $\mathbf{I}_{m_k}$.

\begin{equation}\label{eq:Smatrixexample}
	\begin{bmatrix} Y_{T,t}\\Y_{A,t}\\Y_{B,t}\\Y_{AA,t}\\Y_{AB,t}\\Y_{BA,t}\\Y_{BB,t}\\Y_{BC,t}
	\end{bmatrix} =
	\begin{bmatrix}
		1&1&1&1&1\\1&1&0&0&0\\0&0&1&1&1\\1&0&0&0&0\\0&1&0&0&0\\0&0&1&0&0\\0&0&0&1&0\\0&0&0&0&1\\
	\end{bmatrix}
	\begin{bmatrix}
		Y_{AA,t}\\Y_{AB,t}\\Y_{BA,t}\\Y_{BB,t}\\Y_{BC,t}
	\end{bmatrix}.
\end{equation}

In general, given $h$-step-ahead independent forecasts for the base series $\hat{\mathbf{Y}}_n(h)$, a final forecast $\tilde{\mathbf{Y}}_n(h)$ can be estimated for all series as shown in (\ref{eq:unified_format}), where $\mathbf{G}$ is an $m_k \times m$ matrix.

\begin{equation}
	\label{eq:unified_format}
	\tilde{\mathbf{Y}}_n(h) = \mathbf{S}\mathbf{G}\hat{\mathbf{Y}}_n(h)
\end{equation}

\subsection{Forecast combination}	\label{sec:combination}

Forecast combination methods~\cite{Bates1969} typically forecast the time series of one hierarchy level, and extrapolate the forecasts via aggregation or proportion. In this way, the coherence of the forecasts is guaranteed.

\subsubsection{Bottom-up (BU)}

It is arguably the simplest hierarchical forecast method, consisting of forecasting only the base level of the hierarchy, and then aggregating the forecasts using $\mathbf{S}$ for the rest of the hierarchy. The matrix $\mathbf{G}$ is defined as shown in (\ref{eq:G_BU}), which yields $\mathbf{S}\mathbf{G_{BU}} = \mathbf{S}$.

\begin{equation}
	\label{eq:G_BU}
	\mathbf{G}_{BU} =
	\begin{bmatrix}
		\mathbf{0}_{m_k \times (m-m_k)} & \mathbf{I}_{m_k}
	\end{bmatrix}
\end{equation}

\subsubsection{Top-down (TD)}

It computes forecasts only for the top level of the hierarchy, and then splits the values across all hierarchical levels using factors, which are estimated from previous values of the series. Typically, the proportions of historical averages $\mathbf{p}$ are used~\cite{Athanasopoulos2020,Abolghasemi2022}, as shown in (\ref{eq:p_TD}), in combination with matrix  $\mathbf{G}$ (\ref{eq:G_TD}). A drawback of this approach is that the forecasts are biased, for any $\mathbf{p}$, even if the base forecasts are unbiased~\cite{Hyndman2011}.

\begin{equation}
	\label{eq:p_TD}
	p_j = \frac{\sum_{t=1}^n Y_{j,t}}{\sum_{t=1}^n Y_{t}}, \quad j=1, \dots, m_k
\end{equation}

\begin{equation}
	\label{eq:G_TD}
	\mathbf{G}_{TD} =
	\begin{bmatrix}
		\mathbf{p} & \mathbf{0}_{m_k \times (m-1)}
	\end{bmatrix}
\end{equation}

\subsection{Forecast reconciliation}	\label{sec:reconciliation}

Forecast reconciliation methods, although sometimes categorized as combination methods, have the particularity that a forecast is computed for every individual time series in the hierarchy, and these forecasts are then adjusted to increase their coherence.

\subsubsection{Generalized Least Squares (GLS)}
The base forecasts can be rewritten as in~(\ref{eq:base_forecast}), where $\boldsymbol{\beta}_n(h)$ is the unknown mean of the future values of the bottom level of the hierarchy, and $\boldsymbol{\epsilon}_h$ has zero mean and covariance matrix $\mathbf{\Sigma}_h$. Additionally, Hyndman et al.~\cite{Hyndman2011} demonstrated that adjusted forecasts (following~(\ref{eq:unified_format})) are unbiased if and only if constraint~(\ref{eq:unbiased_constraint}) is met.

\begin{equation}
	\label{eq:base_forecast}
	\begin{split}
		\hat{\mathbf{Y}}_n(h) &= \mathbf{S}\boldsymbol{\beta}_n(h) + \boldsymbol{\epsilon}_h \\
		\boldsymbol{\beta}_n(h) &= E[\mathbf{Y}_{k,n+h} | \mathbf{Y}_1, \dots, \mathbf{Y}_{m_k}]
	\end{split}
\end{equation}

\begin{equation}
	\label{eq:unbiased_constraint}
	\mathbf{S}\mathbf{G}\mathbf{S}=\mathbf{S}
\end{equation}

Combining these definitions, an unbiased estimate for $\boldsymbol{\beta}_n(h)$ can be obtained by GLS estimation as in~(\ref{eq:regression}), where $\mathbf{\Sigma}_h^\dagger$ is the Moore-Penrose pseudoinverse of $\mathbf{\Sigma}_h$. The main difficulty with this formulation is that estimating $\mathbf{\Sigma}_h$ and subsequently computing $\mathbf{\Sigma}_h^\dagger$ can be difficult for large hierarchies. The different reconciliation methods described in the next few subsections use different estimates for $\mathbf{\Sigma}_h$.

\begin{equation}
	\label{eq:regression}
	\begin{split}
		\hat{\boldsymbol{\beta}}_n(h) &= (\mathbf{S}' \mathbf{\Sigma}_h^\dagger  \mathbf{S})^{-1}\mathbf{S}' \mathbf{\Sigma}_h^\dagger \hat{\mathbf{Y}}_n(h) \\
		\tilde{\mathbf{Y}}_n(h) &= \mathbf{S}\hat{\boldsymbol{\beta}}_n(h) = \mathbf{S}\mathbf{G}_{GLS}\hat{\mathbf{Y}}_n(h) \\
		\mathbf{G}_{GLS} &= (\mathbf{S}' \mathbf{\Sigma}_h^\dagger  \mathbf{S})^{-1}\mathbf{S}' \mathbf{\Sigma}_h^\dagger
	\end{split}
\end{equation}

\subsubsection{Ordinary Least Squares (OLS)}
By assuming that the errors are aggregated across the hierarchy in the same way as the series themselves ($\boldsymbol{\epsilon}_h \approx \mathbf{S}\boldsymbol{\epsilon}_{k,h}$), OLS can be applied and $\mathbf{G}$ can be derived as shown in~(\ref{eq:ols})~\cite{Hyndman2011}. 
Interestingly, this optimal value for $\mathbf{G}$ depends exclusively on the summing matrix and not on the data itself.

\begin{equation}
	\label{eq:ols}
	\mathbf{G}_{OLS} = \mathbf{S}(\mathbf{S}'\mathbf{S})^{-1}\mathbf{S}'
\end{equation}

\subsubsection{Minimum Trace (MinT)}

Wickramasuriya et al.~\cite{Wickramasuriya2019} showed that the covariance matrix of the coherent forecast errors follows~(\ref{eq:variance_coherent_forecast_errors}), where $\mathbf{W}_h$ is the covariance matrix of the base forecast errors. In this setting, we want to find the $\mathbf{G}$ that minimizes the error variances, whose sum is the trace of $\mathbf{V}_h$, while respecting the constraint for an unbiased forecast in~(\ref{eq:unbiased_constraint}), as shown in~(\ref{eq:mint}).

\begin{equation}
	\label{eq:variance_coherent_forecast_errors}
	\begin{split}
		\mathbf{V}_h &= \mathrm{Var} \left[ \mathbf{Y}_{n+h} - \tilde{\mathbf{Y}}_n(h)\right] = \mathbf{S}\mathbf{G}\mathbf{W}_h\mathbf{G}'\mathbf{S}' \\
		\mathbf{W}_h &= \mathrm{Var} \left[ \mathbf{Y}_{n+h} - \hat{\mathbf{Y}}_n(h)\right]
	\end{split}
\end{equation}

\begin{equation}
	\label{eq:mint}
	\begin{split}
		\mathbf{G}_{MinT} &= (\mathbf{S}'\mathbf{W}^{-1}_h\mathbf{S})^{-1}\mathbf{S}'\mathbf{W}^{-1}_h \\
		\tilde{\mathbf{Y}}_n(h) &= \mathbf{S}(\mathbf{S}'\mathbf{W}^{-1}_h\mathbf{S})^{-1}\mathbf{S}'\mathbf{W}^{-1}_h\hat{\mathbf{Y}}_n(h)
	\end{split}
\end{equation}

Naturally, $\mathbf{W}_h$ is unknown and has to be estimated from the training data; different estimates yield different reconciliation methods. Note that assuming that $\mathbf{G}$ is independent of the data---and therefore $\mathbf{W}_h = \mathbf{I_m}$---yields OLS. Also, note that the estimates of $\mathbf{W}_h$ usually include a scaling factor in the original publications~\cite{Wickramasuriya2019,Hyndman2021}, which is only relevant when computing confidence intervals and is canceled out when applying~(\ref{eq:mint}). For the sake of simplicity, in this paper we assume that the constant is $1$.

Equation~(\ref{eq:mint_w}) shows a more relaxed assumption: that the error covariance matrices are proportional to each other. This allows us to estimate the matrix for $h=1$, and derive $\mathbf{W}_h$ from it. In this case, $\mathbf{W}_1$ is estimated based on the sample covariance of the residuals. Throughout this paper, this is the method we will refer to as \textit{MinT}.

\begin{equation}
	\label{eq:mint_w}
	\mathbf{W}_h = \mathbf{W}_1 
\end{equation}

\subsubsection{Weighted Least Squares with Variance scaling (WLSV)}

When $\mathbf{W}_h$ is defined as in~(\ref{eq:WLSV}), where $\mathbf{e}_t$ is an $m$-dimensional vector containing the residuals of the base models trained on each series, the weights for each series are scaled according to the variance of the resisuals of the base forecasts.

\begin{equation}
	\label{eq:WLSV}
	\begin{split}
		\mathbf{W}_h &= \mathrm{diag}(\hat{\mathbf{W}}_1) \\
		\hat{\mathbf{W}}_1 &= \frac{1}{T} \sum^T_{t=1}{\mathbf{e}_t\mathbf{e}'_t} \\
	\end{split}
\end{equation}

\subsubsection{Weighted Least Squares with Structural scaling (WLSS)}

In this case, we define $\mathbf{W}_h$ as in Equation~\ref{eq:WLSS}. In other words, each value along the diagonal of $\mathbf{W}_h$ corresponds to the sum of the corresponding row in $\mathbf{S}$, which makes it dependent exclusively on the structure of the hierarchy and not on the values of the time series nor the models.

\begin{equation}
	\label{eq:WLSS}
	\mathbf{W}_h = \mathrm{diag}(\mathbf{S}\mathbf{1}_{m_k}) 
\end{equation}

\subsubsection{Non-negative reconciliation (NNOLS and NNMinT)}
Previously described methods do not ensure that the reconciled forecasts are non-negative, which is a constraint in many real-life problems. In fact, a sufficient condition for the non-negativity of $\tilde{\mathbf{Y}}_n(h)$ is that vectors $\tilde{\mathbf{b}}_n(h)=\mathbf{G}\hat{\mathbf{Y}}_n(h)$ should be non-negative, as can be derived from~(\ref{eq:unified_format}). In~\cite{Wickramasuriya2020}, it is demonstrated that this the optimal $\tilde{\mathbf{b}}_n(h)$ is the solution of the quadratic programming problem in~(\ref{eq:nnmint}).

\begin{equation}
	\label{eq:nnmint}
	\begin{split}
		\tilde{\mathbf{b}}_n(h) = \min_{\mathring{\mathbf{b}}}&\frac{1}{2} \mathring{\mathbf{b}}'\mathbf{S}'\mathbf{W}^{-1}_h\mathbf{S}\mathring{\mathbf{b}}-\mathring{\mathbf{b}}'\mathbf{S}'\mathbf{W}^{-1}_h\hat{\mathbf{Y}}_n(h) \\
		& s.t. ~ \mathring{b}_i \ge 0 \quad \forall i \in \{1,\dots,m\}
	\end{split}
\end{equation}

\section{Forecast reconciliation of vaccine sales}	\label{sec:proposal}

In this paper, we applied classic and state of the art methods to reconcile vaccine sales hierarchical time series. Section~\ref{sec:dataset} describes the dataset tackled. Then, Section~\ref{sec:preprocessing} describes the preprocessing applied on the dataset, Section~\ref{sec:methods} depicts the combination and reconciliation methods that were considered in this study, and Section~\ref{sec:model_selection} describes the model selection prodecure followed to choose the hyperparameters for the forecast.

\subsection{Dataset}	\label{sec:dataset}
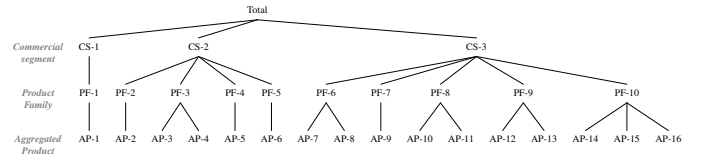
\begin{figure}
	\centering
	\begin{center}
		\resizebox{0.5 \textwidth}{!}{
			\begin{forest} 
				for tree={
					scale=.75, sibling distance=0pt,
					grow=south, 
					edge={line cap=round},
					outer sep=0pt, 
					rounded corners,
					minimum height=1mm 
				}
				[Total 
				[\textit{\textbf{Commercial}} \\ \textit{\textbf{segment}}, for tree={color=gray,no edge}  
				[\textit{\textbf{Product}} \\ \textit{\textbf{Family}}, tier=b                            
				[\textit{\textbf{Aggregated}} \\ \textit{\textbf{Product}},  tier=c]]]
				[CS-1 
				[PF-1, tier=b
				[AP-1, tier=c]]]
				[CS-2 
				[PF-2, tier=b
				[AP-2, tier=c]]
				[PF-3, tier=b
				[AP-3, tier=c] [AP-4, tier=c]] 
				[PF-4, tier=b
				[AP-5, tier=c]] 
				[PF-5, tier=b
				[AP-6, tier=c]]]
				[CS-3
				[PF-6, tier=b
				[AP-7, tier=c]
				[AP-8, tier=c]]
				[PF-7, tier=b
				[AP-9, tier=c]]
				[PF-8, tier=b
				[AP-10, tier=c]
				[AP-11, tier=c]]
				[PF-9, tier=b
				[AP-12, tier=c]
				[AP-13, tier=c]]
				[PF-10, tier=b
				[AP-14, tier=c]
				[AP-15, tier=c]
				[AP-16, tier=c]]]
				]
				\label{hier2}%
			\end{forest}
		}
	\end{center}
	\caption{Hierarchical representation of the dataset}
	\label{fig:hierarchy}
\end{figure}

The dataset used for this study consists of monthly vaccine sales from GSK, between years 2010 and 2021. The time series hierarchy has 5 levels: commercial segment, product family, aggregated product, product type and stock keeping unit. To ensure that the lowest level time series contain sufficient data, we focus on the top 3 levels of the hierarchy (noted as CS, PF and AP respectively), which are composed by 3, 10 and 16 time series, respectively, as shown in Fig.~\ref{fig:hierarchy}.
It is noteworthy that some series start later than 2010, typically because they refer to recently developed products.

Because the time series represent sales numbers, their values are, in general, non-negative integers. However, the raw time series collected directly from the sales department also contains corrections done at the end of the year to compensate for registration errors and ensure that yearly totals are correct. This causes some of the time series values to have unexpectedly large or small values, leading to negative values and breaking seasonal patterns, which in turns adds to the complexity of the problem.

The data was split into training and forecast periods. Due to a potentially different behavior of the time series during the COVID-19 pandemic, two different splits were considered, which either excluded the pandemic period from the data, or included it entirely in the forecast period, as shown in Table~\ref{tab:dataset_partitioning}.

\begin{table}
	\centering
	
	\caption{Train and forecast period splits of data.}
	\label{tab:dataset_partitioning}
	\begin{tabular}{lll}
		\hline
		Period                  & Before COVID-19  & After COVID-19 \\
		\hline
		Train period            & 2010 - 2017      & 2010 - 2019 \\
		Forecast period         & 2018 - 2019      & 2020 - 2021 \\
		\hline
	\end{tabular}
\end{table}

\subsection{Preprocessing}	\label{sec:preprocessing}

Negative values in the time series were replaced by a linear interpolation of the preceding and following values, while shifting all the values for that year proportionally, to ensure that the total sales volume for the year remains the same. This was done at the PF level of the hierarchy.

\begin{figure}
	\centering
	\begin{subfigure}[b]{0.5\textwidth}
		\centering
		\includegraphics[width=1.\textwidth]{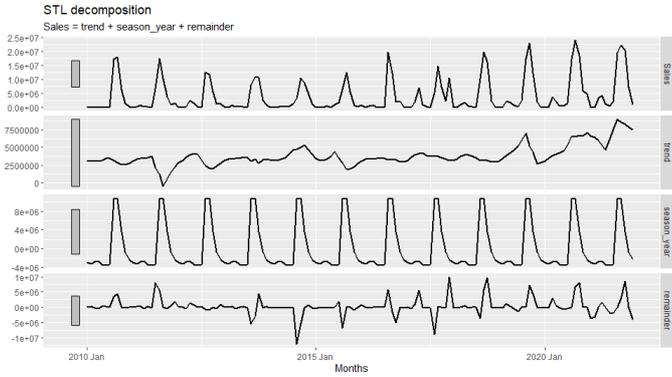}%
		\caption{STL decomposition of PF3. Outlier detection is made on the remainder component.}%
		\label{fig:STL}%
	\end{subfigure} \\
	\begin{subfigure}[b]{0.5\textwidth}
		\centering
		\includegraphics[width=1.\textwidth]{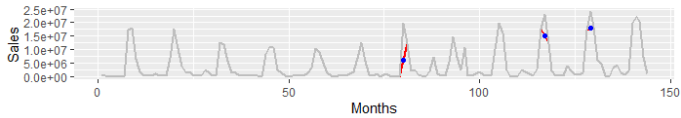}%
		\caption{Detected outliers from the remainder of above decomposition are replaced by linear interpolation. The blue dot is the replacement for the outlier, and the red line is the smoothed curve.}%
		\label{fig:outlier}%
	\end{subfigure}
	\caption{STL decomposition and outlier detection on the remainder.}
	\label{fig:stl_outlier}
\end{figure}

The outlier detection was carried out by first applying a Seasonal and Trend decomposition using Loess decomposition (STL)~\cite{Cleveland1990}, to extract seasonal, trend and remainder components from the series. Then, values in the remainder that deviate from the quartiles by more than 3 times the inter-quartile range were considered to be outliers, following other works in the field~\cite{Hyndman2021}. As an example, Fig.~\ref{fig:stl_outlier} shows the STL decomposition and the corrected outliers of time series PF3, which is clearly seasonal. 

\subsection{Forecasting}	\label{sec:methods}

\begin{figure}[hbt!]%
	\centering
	\includegraphics[width=0.5\textwidth]{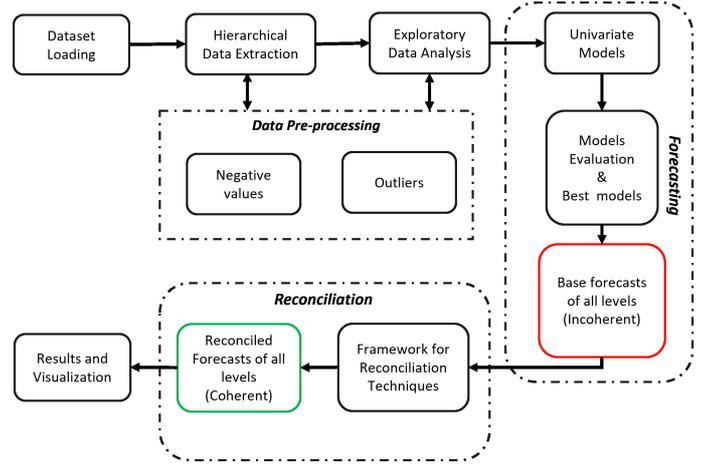}%
	\caption{General workflow of the proposal.}%
	\label{fig:workflow}%
\end{figure}

The entire pipeline applied on this problem is shown on Fig.~\ref{fig:workflow}.

The time series were normalized with a Box-Cox transformation (Equation~\ref{eq:boxcox}), with $\lambda$ being estimated using profile likelihood function and goodness of fit tests for normality, and where $\lambda_2=\min{y_i}$.

\begin{equation}
	\label{eq:boxcox}
	y_{i}^{(\lambda)}= \begin{cases}
		\frac{(y_{i} + \lambda_2) ^{\lambda}-1}{\lambda} &
		\text { if } \lambda \neq 0 \\ \ln (y_{i}+ \lambda_2) &
		\text { if } \lambda=0
	\end{cases}
\end{equation}

After normalization, we train a seasonal ARIMA model for each time series in the hierarchy. Naturally, each of these time series might have a different seasonality and dependencies on lagged values; therefore, a model selection has to be done when training each series.

\subsection{Model selection}	\label{sec:model_selection}
To compute the baseline forecasts of the time series, Seasonal ARIMA~\cite{Hyndman2021} models were trained with 4 different parameters for the seasonality (1, 3, 6 and 12) months. Additionally, the non-seasonal ($p$, $d$ and $q$) and seasonal ($P$, $D$ and $Q$) parameters of ARIMA must be chosen. The differencing term $d$ was estimated using two tests---Kwiatkowski, Phillips, Schmidt–Shin (KPSS) and Dickey–Fuller (ADF)---and taking the maximum value from both. The seasonal differencing $D$ was estimated by Osborn, Chui, Smith, and Birchenhall (OCSB) test. The remaining parameters were estimated via a grid search within the training set, using Akaike's Information Criterion (AIC) as the optimized metric.

\begin{figure}[hbt!]%
	\centering
	\includegraphics[width=0.5\textwidth]{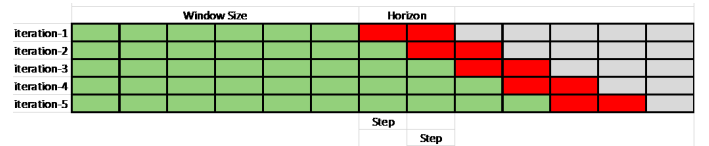}%
	\caption{Cross-validation by rolling origin (training set in green, validation set in red)}%
	\label{fig:rolling_origin}%
\end{figure}

After the 6 hyperparameters for each seasonality have been optimized, we must select the seasonal frequency that obtains the best results. This is done by choosing the one that yields the best symmetric mean absolute percentage error (SMAPE). To this purpose, a specific variant of cross-validation is applied (rolling forecasting origin~\cite{Bergmeir2018}), where the size of the training set is increased for each iteration, and the validation set has a fixed size and always starts immediately after the training set (Fig.~\ref{fig:rolling_origin}).

The rolling forecasting origin was performed with a minimum training set size of 28 months, and a validation set size of 2 months, increasing the size of the training set 1 month at a time. As there are a total of 84 months in the entire training set, this yielded 54 pairs of training and validation sets.

\subsection{Reconciliation}

After computing the forecast for all time series, we reconcile the forecasts using the methods described in Section~\ref{sec:reconciliation}, namely BU, OLS, NNOLS, WLSS, WLSV and MINT.

\subsection{Performance measures}
Two performance measures were used to evaluate the forecasts and their reconciliations:
\begin{description}
	\item[Sales Forecast Bias (SFB)] measures the tendencies to over- and underforecast along a certain period of length $T$, rather than at individual points in time. This measure is often used by companies to evaluate sales predictions, because an over-estimation in a certain month can be compensated by an equivalent under-estimation in the following month (or vice-versa). The SFB of a certain time series $i$ is computed as shown in Equation~\ref{eq:SFB}.
	\begin{equation}
		\label{eq:SFB}
		SFB_i = \frac{\sum_{s=1}^{T} {\left[ \hat{Y}_{i,n-s}(h) - Y_{i,n-s}(h) \right]} }{\sum_{s=1}^{T} Y_{i,n-s}(h)} \times 100
	\end{equation}
	In this work, we calculated the SFB as a percentage over a period of 6 months ($T=6$). To aggregate results for multiple time series while avoiding the cancellation of positive and negative errors, we computed the weighted average of the absolute values of SFB for all series (volume-Weighted SFB, WSFB, Equation~\ref{eq:WSFB}), where the weight $w_i$ is the sum of all values of a time series:
	
	\begin{equation}
		\label{eq:WSFB}
		\begin{split}
			WSFB & = \frac{\sum_{i=1}^{m}{w_i SFB_i}}{\sum_{i=1}^{m}{w_i}} \\
			w_i & = \sum{Y_{i,t}}
		\end{split}
	\end{equation}

	\item[Root mean squared scaled error (RMSSE)] is an error measure scaled by the difference between consecutive values in the series as shown in Equation~\ref{eq:RMSSE}.
	\begin{equation}
		\label{eq:RMSSE}
		RMSSE = \sqrt{\frac{\frac{1}{h} \sum_{t=n+1}^{n+h}(y_t - \hat{y}_t)^2 }
			{\frac{1}{n-1} \sum_{t=2}^{n}(y_t - y_{t-1})^2 }}
	\end{equation}
	
	Similarly to the SFB in Equation~\ref{eq:WSFB}, we compute the weighted average of the RMSSE when aggregating over multiple time series.
\end{description}

\section{Results}	\label{sec:results}

This section describes the results obtained from the experiments. First, baseline results are presented in Section~\ref{sec:results_baseline}. The results of the different reconciliation methods are shown in Section~\ref{sec:results_reconciliation}. Section~\ref{sec:results_covid} compares the behavior of the results, and finally Section~\ref{sec:discussion} presents a general discussion of the results.

\subsection{Baseline forecasting}	\label{sec:results_baseline}

\begin{figure}[h]
	\centering
	\includegraphics[width=0.5\textwidth]{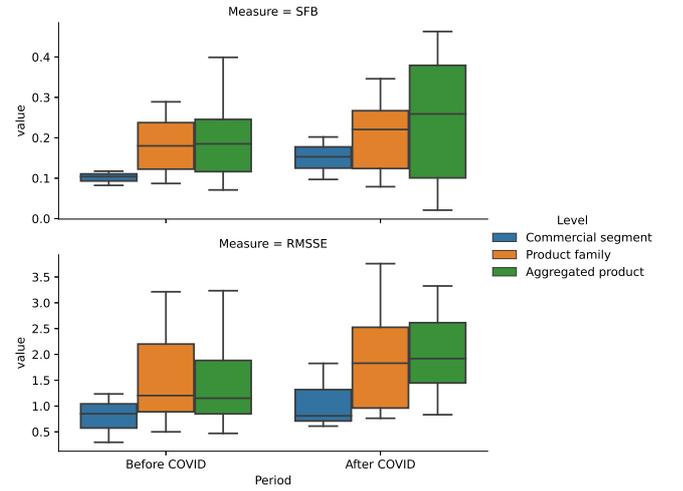}
	\caption{Distribution of baseline results} 
	\label{fig:baseline_boxplot}
\end{figure}

Table~\ref{tab:baseline} and Fig.~\ref{fig:baseline_boxplot} present an overview of the SFB and RMSSE obtained using the baseline forecast method for each series, at each level of the hierarchy. As expected, it can clearly be seen that the highest level in the hierarchy (Commercial segment) is easier to predict than lower levels, whose series are subject to a higher variability. However, the difference in forecast error between the two lowest levels is very small, and the range of obtained errors differs greatly. Some series are very accurately predicted, while others have a large forecast error. The plots also reflect an increase in the errors after COVID-19.

\begin{table}
	\centering
	\caption{Baseline ARIMA results for each hierarchical level} 
	\label{tab:baseline}
	\resizebox{0.5\textwidth}{!}{%
		\begin{tabular}{ll|rr|rr}
			\hline
			&  & \multicolumn{2}{c|}{SFB} & \multicolumn{2}{c}{RMSSE} \\
			&  & Before COVID & After COVID & Before COVID & After COVID \\
			Level & Series &  &  &  &  \\
			\hline
			\multirow[c]{3}{*}{Commercial segment} & CS1 & 0.0823 & 0.0970 & 0.8500 & 0.8101 \\
			& CS2 & 0.1040 & 0.1530 & 0.2975 & 0.6108 \\
			& CS3 & 0.1174 & 0.2020 & 1.2340 & 1.8258 \\
			\hline
			\multirow[c]{10}{*}{Product family} & PF1 & 0.1200 & 0.1330 & 1.2514 & 1.2734 \\
			& PF2 & 0.1400 & 0.0790 & 0.9094 & 0.8255 \\
			& PF3 & 0.2200 & 0.2710 & 1.7796 & 2.5599 \\
			& PF4 & 0.0870 & 0.2550 & 0.8804 & 2.1496 \\
			& PF5 & 0.2300 & 0.2540 & 3.2135 & 2.7705 \\
			& PF6 & 0.2700 & 0.1210 & 1.1544 & 1.5091 \\
			& PF7 & 0.2890 & 0.3460 & 0.5860 & 0.7630 \\
			& PF8 & 0.1110 & 0.0970 & 0.4999 & 0.8568 \\
			& PF9 & 0.1300 & 0.1870 & 2.3414 & 2.4231 \\
			& PF10 & 0.2400 & 0.2820 & 2.6340 & 3.7584 \\
			\hline
			\multirow[c]{16}{*}{Aggregated product} & AP1 & 0.0860 & 0.1810 & 1.7226 & 2.4580 \\
			& AP2 & 0.1020 & 0.0410 & 0.8710 & 0.8318 \\
			& AP3 & 0.1940 & 0.1710 & 0.7796 & 1.1590 \\
			& AP4 & 0.1870 & 0.2910 & 2.5067 & 3.0681 \\
			& AP5 & 0.2960 & 0.2760 & 3.2135 & 3.0005 \\
			& AP6 & 0.1210 & 0.1010 & 0.6460 & 0.8516 \\
			& AP7 & 0.1290 & 0.3980 & 1.1353 & 1.2203 \\
			& AP8 & 0.1830 & 0.0210 & 1.1544 & 2.5091 \\
			& AP9 & 0.3670 & 0.4120 & 1.1474 & 2.9291 \\
			& AP10 & 0.2440 & 0.4630 & 0.4702 & 1.8843 \\
			& AP11 & 0.0750 & 0.1000 & 1.1219 & 1.9539 \\
			& AP12 & 0.0710 & 0.0970 & 0.4999 & 1.8357 \\
			& AP13 & 0.3990 & 0.3830 & 1.8604 & 1.5216 \\
			& AP14 & 0.2320 & 0.2420 & 1.9526 & 2.3568 \\
			& AP15 & 0.1460 & 0.2820 & 3.2340 & 3.3258 \\
			& AP16 & 0.2500 & 0.3780 & 1.1624 & 1.7590 \\
			\hline
		\end{tabular}
	}
\end{table}

\subsection{Reconciliation}	\label{sec:results_reconciliation}

\begin{figure*}[h]
	\centering
	\includegraphics[width=\textwidth]{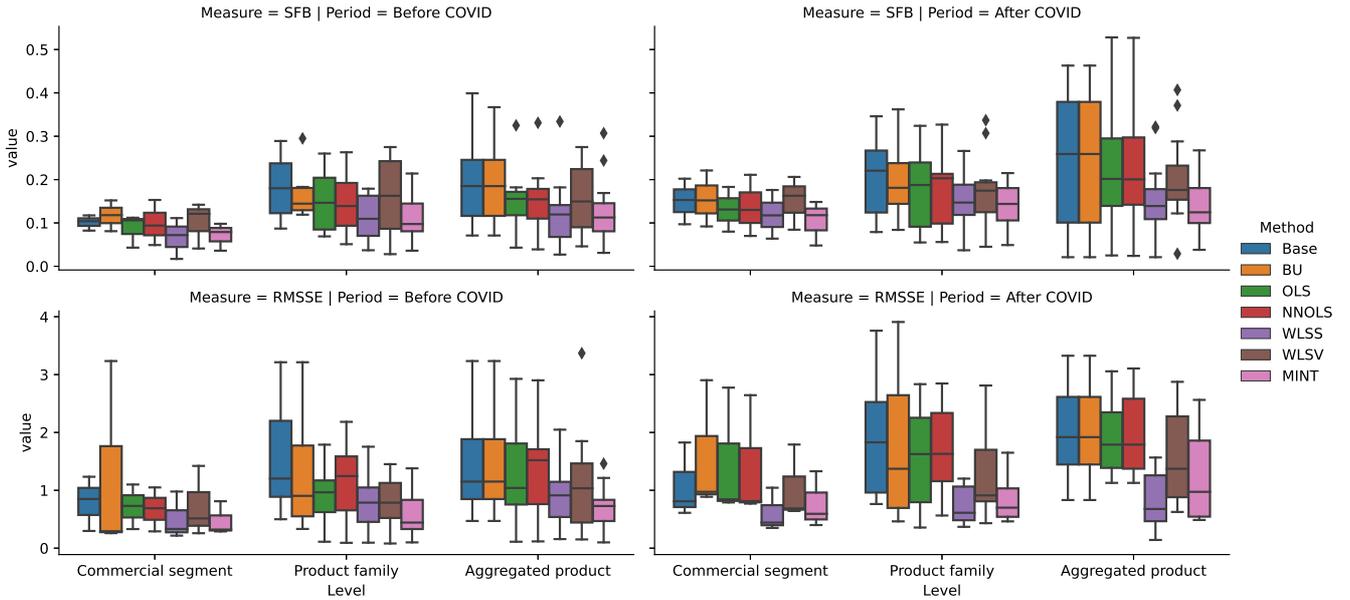}
	\caption{Distribution of results after reconciliation} 
	\label{fig:reconciliation_boxplot}
\end{figure*}

Figure~\ref{fig:reconciliation_boxplot} shows the SFB and RMSSE obtained for all time series and reconciliation methods. As was observed in the previous section, we can see that the highest level of the hierarchy is easier to predict than the two lowest levels, a behavior that is consistent across all methods tested.

\begin{figure*}[h]
	\centering
	\includegraphics[width=\textwidth]{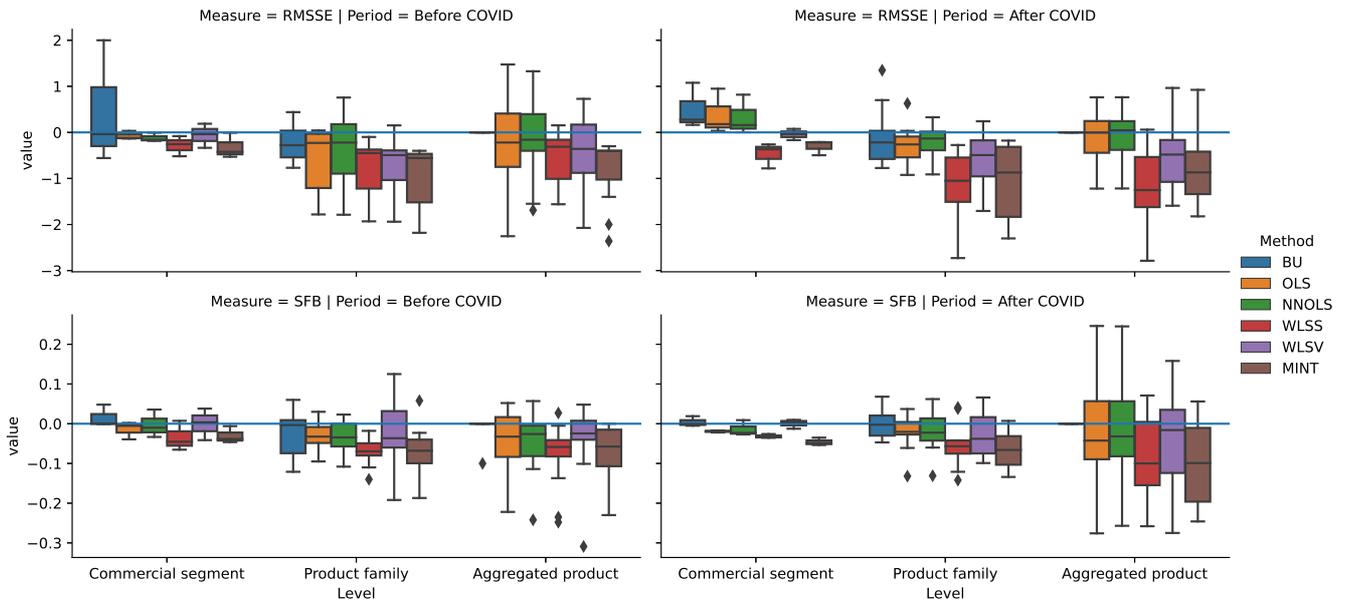}
	\caption{Distribution of the differences with respect to base forecast.} 
	\label{fig:reconciliation_diff_boxplot}
\end{figure*}

Although all methods seem to improve on the baseline, MINT and WLSS obtain the lowest SRB and RMSSE in the two tested periods and for all hierarchy levels. However, the boxplots show a large overlap of the error values, which makes it difficult to conclude whether the results of all individual time series are consistently improved by these methods. Figure~\ref{fig:reconciliation_diff_boxplot} shows the difference in SFB and RMSSE obtained by each method with respect to the result obtained by the baseline. We also see that most differences are indeed negative, indicating that in a vast majority of the cases the results of the baseline are improved. As expected, the combination method BU doesn't modify the forecast for the base level of thei hierarchy. WLSS and MINT improve the results almost in all cases.

For reference, the complete results are reported in the Tables~\ref{tab:reconciliation_sfb} and~\ref{tab:reconciliation_rmsse}.

%
%

	\subsection{Comparison before/after COVID-19}	\label{sec:results_covid}
	In the previous figures, it could be seen that, in general, the error increased in the period after COVID-19. As initially expected, forecasting the behavior of the series after the start COVID-19 is more difficult, as the test data differs more from the training data. This trend is observed for all methods	and levels, but is more accused in the lowest level of the hierarchy, where we can see that some forecasts achieve very large errors, and also that, for some of the series, the reconciliation methods produce forecasts that are worse than the baseline. Only WLSS and MINT seem to escape this trend, having the 75th percentile of the difference in SFB and RMSSE (Fig.~\ref{fig:reconciliation_diff_boxplot}) below zero for both measures and for all hierarchy levels, indicating that at least 75\% of the time series have their forecast improved by these reconciliation methods.

	
	\subsection{Discussion}	\label{sec:discussion}
	
	In this section, we carry out a statistical analysis of the results obtained in this experimental study, in order to determine if the differences observed in the results of the different methods are statistically significant or not. First, Table~\ref{tab:friedman} shows the results of the Friedman test to determine whether there is any difference in the distributions of the SFB and RMSSE values obtained by all methods.
	
	\begin{table}
		\centering
		\caption{Results of Friedman test}
		\label{tab:friedman}
		\begin{tabular}{lrr}
			\hline
			& Statistic & p-value \\
			\hline
			SFB & 108.1 & 5.081e-21 \\
			RMSSE & 164.2 & 7.504e-33 \\
			\hline
		\end{tabular}
	\end{table}
	
	Clearly, there are differences in the distributions, which enables us to apply a Dunn test to evaluate potential differences in the results of every combination and reconciliation method, using the baseline as a reference in every case. Holm's method was applied for p-value adjustment. The results of this test are shown in Table~\ref{tab:dunn}. In the table, we can see that WLSS and MINT obtain statistically significant differences with the baseline method. This result is in line with the previous observation that these two algorithms outperformed the rest systematically, for all aggregation levels, both before and after COVID.
	
	To conclude the study, Table~\ref{tab:wilcoxon} shows the results of Wilcoxon's test to compare the performance of all methods, at all levels, before and after COVID. Clearly, there is a statistically significant performance drop when test values after COVID are considered, despite the use of a larger training set, indicating an important change in the behavior of the time series analyzed.

	\begin{table}
		\centering
		\caption{Results of Dunn test with Holm p-value adjustment}
		\label{tab:dunn}
		\begin{tabular}{lrr}
			\hline
			& SFB & RMSSE \\
			\hline
			Base & 1 & 1 \\
			BU & 1 & 1 \\
			OLS & 1 & 1 \\
			NNOLS & 1 & 1 \\
			WLSS & 0.004067 & 1.012e-06 \\
			WLSV & 1 & 0.06092 \\
			MINT & 0.001384 & 1.53e-06 \\
			
			\hline
		\end{tabular}
	\end{table}
	
	\begin{table}
		\centering
		\caption{Results of Wilcoxon's test with Bonferroni correction}
		\label{tab:wilcoxon}
		\begin{tabular}{lrr}
			\hline
			& Statistic & p-value \\
			\hline
			SFB & 4546 & 9.988e-11 \\
			RMSSE & 3488 & 5.116e-16 \\
			
			\hline
		\end{tabular}
	\end{table}

	\section{Conclusion}	\label{sec:conclusion}
	
	Vaccine demand prediction is a critical problem in industry, with a large impact on public health and economy. For the first time, we have modeled this problem as a hierarchical time series forecast, establishing a hierarchy based on product families and gathering sales data between 2010 and 2021 in GSK.
	
	Then, we have carried out a comparison of the results of different forecast reconciliation methods for hierarchical time series, applied on a real dataset of vaccine sales containing a 3-level hierarchy over 12 years. The baseline forecast for each level of the hierarchy was computed with ARIMA, and results were computed in a rolled cross-validation scheme.
	
	The analysis has shown that MINT and WLSS obtained the best results for the forecast, a conclusion that is supported by statistical significance o the performed tests, while producing forecasts that are reconciled along the entire hierarchy. Furthermore, we have observed that the algorithms systematically obtain worse forecasts when data during the COVID period is used as test set, despite the use of more years of training data. This indicates a significant change of the behavior of the time series during this period. However, MINT and WLSS are again able to yield better results than the rest of the methods analyzed, palliating the performance drop, indicating their higher robustness to this kind of changes.

	\ack This work was sponsored by GlaxoSmithKline Biologicals SA.
	
	\section*{Author contributions}
	AB, BA, DP, MD were involved in the conception and design of the study and/or the development of the study protocol. AB, BA participated to the acquisition of data. AB, BA, DP, MD analysed and interpreted the results. All authors were involved in drafting the manuscript or revising it critically for important intellectual content. AB and BA had full access to the data. All authors approved the manuscript before it was submitted by the corresponding author. 
	
	\section*{Conflict of interest}
	MD and AB are employees of the GSK group of companies.
	
	\bibliography{biblio}
	
	\clearpage
	\appendix
	
	\section{Supplementary material: complete result tables}
	\begin{table}
		\centering
		\caption{SFB after reconciliation, for all levels and methods} 
		\label{tab:reconciliation_sfb}
		\resizebox{0.8\textwidth}{!}{%
			\begin{tabular}{l|rrrrrrr|rrrrrrr}
				\hline
				& \multicolumn{7}{c|}{Before COVID} & \multicolumn{7}{c}{After COVID} \\
				Series & Base & BU & OLS & NNOLS & WLSS & WLSV & MINT & Base & BU & OLS & NNOLS & WLSS & WLSV & MINT \\ \hline
				CS1 & .0823 & .0810 & .0430 & .0490 & .0170 & .0410 & .0360 & .0970 & .0920 & .0800 & .0700 & .0638 & .0844 & .0480 \\
				CS2 & .1040 & .1520 & .1060 & .0940 & .1112 & .1420 & .0980 & .1530 & .1520 & .1310 & .1300 & .1175 & .1627 & .1179 \\
				CS3 & .1174 & .1180 & .1120 & .1530 & .0720 & .1211 & .0790 & .2020 & .2210 & .1830 & .2110 & .1760 & .2061 & .1484 \\ \hline
				PF1 & .1200 & .1800 & .0690 & .0710 & .0740 & .1120 & .0800 & .1330 & .1410 & .1650 & .1950 & .1710 & .1980 & .1110 \\
				PF2 & .1400 & .1360 & .0990 & .0980 & .0690 & .0780 & .0360 & .0790 & .1470 & .0880 & .0980 & .0370 & .0450 & .0490 \\
				PF3 & .2200 & .1810 & .1960 & .1930 & .1380 & .0280 & .0880 & .2710 & .2950 & .2100 & .2110 & .1500 & .1720 & .1487 \\
				PF4 & .0870 & .1450 & .0800 & .0920 & .0690 & .1360 & .1450 & .2550 & .2100 & .2260 & .2140 & .1940 & .1810 & .1899 \\
				PF5 & .2300 & .1440 & .2600 & .2450 & .1710 & .2750 & .1440 & .2540 & .2390 & .2440 & .2110 & .1120 & .1770 & .1390 \\
				PF6 & .2700 & .1830 & .1750 & .1620 & .1300 & .2050 & .0830 & .1210 & .1430 & .1010 & .1000 & .0610 & .0790 & .0530 \\
				PF7 & .2890 & .2950 & .2070 & .1900 & .1790 & .2670 & .2140 & .3460 & .3620 & .3240 & .3220 & .2660 & .3370 & .2120 \\
				PF8 & .1110 & .1210 & .0700 & .0510 & .0370 & .0570 & .0710 & .0970 & .0840 & .0760 & .0940 & .1380 & .1630 & .1040 \\
				PF9 & .1300 & .1270 & .1180 & .1160 & .0890 & .2550 & .1070 & .1870 & .1520 & .0550 & .0560 & .1440 & .1120 & .1520 \\
				PF10 & .2400 & .1190 & .2520 & .2630 & .1720 & .1890 & .1790 & .2820 & .2350 & .3190 & .3268 & .2287 & .3070 & .2150 \\ \hline
				AP1 & .0860 & .0860 & .0430 & .0390 & .0300 & .0460 & .0310 & .1810 & .1810 & .1100 & .1080 & .0870 & .1610 & .1050 \\
				AP2 & .1020 & .1020 & .1200 & .1040 & .0410 & .0730 & .0820 & .0410 & .0410 & .0250 & .0240 & .0210 & .0290 & .0380 \\
				AP3 & .1940 & .1940 & .1570 & .1780 & .1330 & .2400 & .0880 & .1710 & .1710 & .2050 & .2090 & .1130 & .2180 & .0800 \\
				AP4 & .1870 & .1870 & .1620 & .1660 & .1380 & .0900 & .0770 & .2910 & .2910 & .2980 & .3240 & .1470 & .1670 & .1090 \\
				AP5 & .2960 & .2960 & .1700 & .1950 & .0610 & .2750 & .1440 & .2760 & .2760 & .1700 & .1950 & .1610 & .2750 & .1140 \\
				AP6 & .1210 & .1210 & .1110 & .1120 & .1150 & .1260 & .1210 & .1010 & .1010 & .1980 & .1970 & .1720 & .1220 & .1570 \\
				AP7 & .1290 & .1290 & .1810 & .1800 & .1560 & .1440 & .0990 & .3980 & .3980 & .2590 & .2590 & .1400 & .1230 & .1520 \\
				AP8 & .1830 & .1830 & .1550 & .1520 & .1300 & .1550 & .1130 & .0210 & .0210 & .1010 & .1000 & .0510 & .1790 & .0430 \\
				AP9 & .3670 & .3670 & .3250 & .3310 & .3340 & .2660 & .3070 & .4120 & .4120 & .3420 & .3650 & .2140 & .2880 & .1960 \\
				AP10 & .2440 & .2440 & .1370 & .1300 & .1820 & .2410 & .2440 & .4630 & .4630 & .2940 & .2970 & .3220 & .3710 & .2674 \\
				AP11 & .0750 & .0750 & .0912 & .0680 & .0700 & .1230 & .0750 & .1000 & .1000 & .1490 & .1490 & .0960 & .1310 & .1080 \\
				AP12 & .0710 & .0710 & .1000 & .0810 & .0270 & .0570 & .0710 & .0970 & .0970 & .2060 & .2040 & .1380 & .1630 & .0840 \\
				AP13 & .3990 & .2990 & .1770 & .1570 & .1510 & .0900 & .1690 & .3830 & .3830 & .2990 & .2980 & .1960 & .1730 & .1820 \\
				AP14 & .2320 & .2320 & .1520 & .1510 & .1240 & .1910 & .1120 & .2420 & .2420 & .1730 & .1710 & .1360 & .2120 & .1350 \\
				AP15 & .1460 & .1460 & .1820 & .2030 & .0720 & .1890 & .1190 & .2820 & .2820 & .5280 & .5270 & .3190 & .4070 & .2350 \\
				AP16 & .2500 & .2500 & .1560 & .1670 & .1130 & .2190 & .1500 & .3780 & .3780 & .1020 & .1210 & .1330 & .1970 & .1800 \\ \hline
			\end{tabular}
		}
	\end{table}

	\begin{table}
		\centering
		\caption{RMSSE after reconciliation, for all levels and methods} 
		\label{tab:reconciliation_rmsse}
		\resizebox{0.8\textwidth}{!}{%
			\begin{tabular}{l|rrrrrrr|rrrrrrr}
				\hline
				& \multicolumn{7}{c|}{Before COVID} & \multicolumn{7}{c}{After COVID} \\
				Series & Base & BU & OLS & NNOLS & WLSS & WLSV & MINT & Base & BU & OLS & NNOLS & WLSS & WLSV & MINT \\ \hline
				CS1 & .8500 & .2926 & .7300 & .6899 & .3315 & .5148 & .3210 & .8101 & .9735 & .8432 & .8132 & .4421 & .6427 & .5935 \\
				CS2 & .2975 & .2597 & .3300 & .2905 & .2165 & .2585 & .2890 & .6108 & .8876 & .7900 & .7699 & .3496 & .6863 & .3996 \\
				CS3 & 1.2340 & 3.2340 & 1.1000 & 1.0497 & .9785 & 1.4216 & .8110 & 1.8258 & 2.9028 & 2.7746 & 2.6442 & 1.0441 & 1.7909 & 1.3296 \\
				\hline
				PF1 & 1.2514 & .4825 & 1.1825 & 1.1919 & .8311 & .8759 & .3725 & 1.2734 & 1.0944 & .3565 & 1.3551 & .4597 & .9080 & 1.0944 \\
				PF2 & .9094 & .3323 & .9508 & .9544 & .5853 & .4715 & .3163 & .8255 & .6468 & .5642 & .5643 & .3686 & .4318 & .6468 \\
				PF3 & 1.7796 & 1.8292 & 1.7885 & 2.0019 & 1.2811 & .6962 & 1.3796 & 2.5599 & 3.9093 & 2.4894 & 2.4858 & .7114 & .8537 & .5480 \\
				PF4 & .8804 & .4432 & .8115 & 1.6368 & .7807 & 1.0321 & .4042 & 2.1496 & 2.8502 & 1.7114 & 1.7196 & 1.0763 & 2.3910 & .8502 \\
				PF5 & 3.2135 & 3.2135 & 1.7466 & 2.1832 & 1.7532 & 1.2741 & 1.0332 & 2.7705 & 2.0253 & 2.1935 & 2.1563 & 1.1665 & 1.8132 & .7529 \\
				PF6 & 1.1544 & .7544 & 1.1373 & 1.4343 & .7931 & .6726 & .6347 & 1.5091 & .8391 & 1.5415 & 1.5402 & .4812 & .9168 & .5391 \\
				PF7 & .5860 & .8680 & .1549 & .0923 & .1073 & .0806 & .1237 & .7630 & .4630 & 1.3916 & 1.0918 & .4882 & .6541 & .4630 \\
				PF8 & .4999 & .9399 & .1111 & .1169 & .0948 & .1510 & .0999 & .8568 & .5991 & .5954 & .6651 & .5117 & .7951 & .4991 \\
				PF9 & 2.3414 & 1.6080 & .5601 & .5533 & .4098 & 1.4517 & .4802 & 2.4231 & 1.6496 & 2.2739 & 2.3967 & 1.2019 & 1.3616 & 1.6496 \\
				PF10 & 2.6340 & 2.4779 & .9798 & 1.3010 & 1.1216 & 1.1588 & .9013 & 3.7584 & 3.8628 & 2.8346 & 2.8464 & 1.0297 & 2.8109 & 1.4563 \\
				\hline
				AP1 & 1.7226 & 1.7226 & 1.0207 & 1.6699 & .9825 & 1.0449 & .8226 & 2.4580 & 2.4580 & 2.7238 & 2.7099 & 1.2516 & 1.3846 & 1.1081 \\
				AP2 & .8710 & .8710 & .7571 & .7666 & .5767 & .5026 & .4710 & .8318 & .8318 & 1.1283 & 1.1281 & .4200 & .6259 & .4872 \\
				AP3 & .7796 & .7796 & 1.9968 & 1.8277 & .7059 & .9877 & .4796 & 1.1590 & 1.1590 & 1.3443 & 1.3559 & .6043 & 1.7601 & .5480 \\
				AP4 & 2.5067 & 2.5067 & 2.9258 & 2.9010 & 1.4763 & 1.6079 & 1.1067 & 3.0681 & 3.0681 & 1.9185 & 2.8995 & 1.4595 & 2.8748 & 1.8039 \\
				AP5 & 3.2135 & 3.2135 & 1.7466 & 1.6635 & 1.6532 & 3.3697 & 1.2135 & 3.0005 & 3.0005 & 2.1935 & 2.1563 & 1.5665 & 2.4366 & 2.0253 \\
				AP6 & .6460 & .6460 & .4012 & .4044 & .3365 & .2560 & .2460 & .8516 & .8516 & 1.5946 & 1.5943 & .3758 & .6583 & .6329 \\
				AP7 & 1.1353 & 1.1353 & 1.5412 & 1.5380 & 1.0957 & 1.0255 & .7353 & 1.2203 & 1.2203 & 1.4608 & 1.4615 & 1.2796 & 2.1847 & 1.4387 \\
				AP8 & 1.1544 & 1.1544 & 2.1373 & 2.1343 & .9931 & 1.5479 & .7544 & 2.5091 & 2.5091 & 2.5415 & 2.5402 & .4812 & .9168 & .8391 \\
				AP9 & 1.1474 & 1.1474 & 2.6218 & 2.4743 & 1.3002 & 1.4370 & .7474 & 2.9291 & 2.9291 & 3.0567 & 3.1055 & .1419 & 2.8455 & 2.4862 \\
				AP10 & .4702 & .4702 & .1554 & .2553 & .1567 & .1692 & .1002 & 1.8843 & 1.8843 & 1.7840 & 1.6847 & .6024 & .7727 & .5308 \\
				AP11 & 1.1219 & 1.1219 & 1.1889 & 1.1927 & .9676 & 1.8498 & .7219 & 1.9539 & 1.9539 & 1.3540 & 1.3532 & .3079 & 1.0097 & .8235 \\
				AP12 & .4999 & .4999 & .1111 & .1169 & .2948 & .1510 & .0999 & 1.8357 & 1.8357 & 1.7954 & 1.8951 & .6117 & .6951 & .4991 \\
				AP13 & 1.8604 & 1.8604 & .7458 & .7446 & .8582 & .7142 & 1.4604 & 1.5216 & 1.5216 & 2.2831 & 2.2842 & 1.2929 & 2.2325 & 2.4452 \\
				AP14 & 1.9526 & 1.9526 & 1.0600 & 1.5021 & .6333 & 1.0797 & .5526 & 2.3568 & 2.3568 & 1.1362 & 1.1390 & .7425 & 1.2788 & .5336 \\
				AP15 & 3.2340 & 3.2340 & .9798 & 1.5448 & 2.0479 & 1.1599 & .8740 & 3.3258 & 3.3258 & 2.9346 & 2.9464 & 1.0297 & 2.4109 & 2.5628 \\
				AP16 & 1.1624 & 1.1624 & .9668 & .8633 & .4236 & .2642 & .4624 & 1.7590 & 1.7590 & 1.3981 & 1.3792 & .8790 & 1.3590 & 1.2826 \\
				\hline
			\end{tabular}
		}
	\end{table}

\end{document}